\documentclass{article}
\usepackage{spconf,amsmath,graphicx}

\usepackage{multirow}
\usepackage{amssymb}
\usepackage[bookmarks=false]{hyperref}
\usepackage{textcomp}
\usepackage{bbm} 
\usepackage{subcaption}
\usepackage{multirow}
\usepackage{stfloats}
\usepackage{url}
\usepackage{bbm}
\usepackage{color}
\usepackage[numbers,sort&compress]{natbib}

\title{Category-Adaptive Domain Adaptation for Semantic Segmentation \vspace{-6pt}}
\name{Zhiming Wang, Yantian Luo, Danlan Huang, Ning Ge, Jianhua Lu \vspace{-6pt}}
\address{Department of Electronic Engineering, Tsinghua University, Beijing, China\\
				Beijing National Research Center for Information Science and Technology, Beijing, China \vspace{-18pt}}

\begin{document}
    \setlength{\abovedisplayskip}{3pt}
    \setlength{\belowdisplayskip}{3pt}
    \maketitle
    \begin{abstract}
        Unsupervised domain adaptation (UDA) becomes more and more popular in tackling real-world problems without
        ground truth of the target domain. Though tedious annotation
        work is not required, UDA unavoidably faces two problems: 1)
        how to narrow the domain discrepancy to boost the transferring
        performance; 2) how to improve pseudo annotation producing
        mechanism for self-supervised learning (SSL). In this paper,
        we focus on UDA for semantic segmentation task. Firstly, we
        introduce adversarial learning into style gap bridging mechanism
        to keep the style information from two domains in the similar
        space. Secondly, to keep the balance of pseudo labels on each
        category, we propose a category-adaptive threshold mechanism
        to choose category-wise pseudo labels for SSL. The experiments
        are conducted using GTA5 as the source domain, Cityscapes as
        the target domain. The results show that our model outperforms
        the state-of-the-arts with a noticeable gain on cross-domain
        adaptation tasks.
    \end{abstract}
    \begin{keywords}
        unsupervised domain adaptation, semantic segmentation, self-supervised learning
    \end{keywords}
\vspace{-2mm}
\section{Introduction}
\vspace{-2mm}
As a significant task in computer vision, semantic segmentation aims at producing pixel-wise labels for images, and has been widely applied to many different scenes such as auto driving and scene understanding. However, semantic segmentation usually yields unsatisfying performance without enough labeled training samples. What's more, it's very difficult to apply supervised semantic segmentation to the emergent diverse applications since preparing the pixel-wise annotations is time-consuming and expensive. Therefore, supervised learning based methods are unable to meet the requirements of current image segmentation tasks.

\begin{figure}[t]
	\centering
	\includegraphics[width=0.49\textwidth]{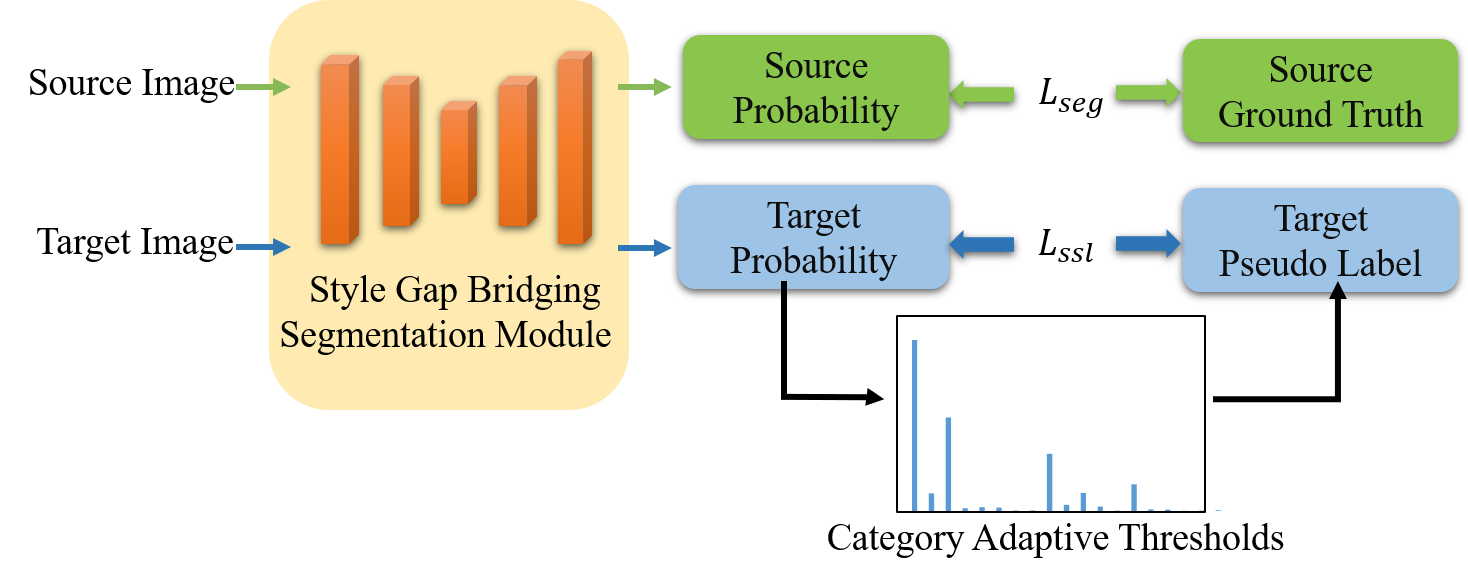}
	\caption{Flowchart of category-adaptive domain adaption approach. Firstly, data from two domains are processed by a style gap bridging mechanism based on adversarial learning, then to boost the performance of SSL, category-adaptive thresholds are adopted to balance the probability of chosen pseudo labels for each category.}
	\label{fig:network_figure}
	\vspace{-4mm} 
\end{figure}

Domain adaptation (DA) offers a solution for semantic segmentation without huge amount of labeled training samples. It aims to apply a model pretrained on the source dataset to generalize on the target dataset. However, there usually exists huge gaps among datasets, which can be categorized into two folds: content-based gap and style-based gap. Content-based gap is caused by inter-dataset amount and frequency discrepancy of categories, which can be alleviated by choosing datasets with similar scenes so that it is often neglected for convenience. The style-based gap refers to the difference of illumination, things' texture and so on. However, modelling the style information is still an open academic problem. It has been illustrated that the shallow layers of CNN extract low-level features (e.g., edges) while the deep layers extract high-level features (e.g., objects) \cite{zeiler2011adaptive}. Based on the fact that different convolutional kernels are computed independently, most literatures regard channel-wise statistics of extracted features as the style information, such as correlation-based Gram matrix \cite{gatys2015neural}, means and standard deviations (evaluated by AdaIN \cite{huang2017arbitrary}). Without loss of generality, in this paper, we adopt the means to model the style information by global average pooling process. However, narrowing the content-based gap is still full of challenges.


Moreover, great advances have been achieved on domain adaptation with SSL, whose key is pseudo labeling mechanism. It solves the problem of lacking available annotations on the target domain. CBST \cite{zou2018unsupervised} introduces the amount of each category as one optimization term so as to balance the probability of pseudo labels of each category. However, each iteration of SSL requires the ordering operation, which is time-consuming. BDL \cite{li2019bidirectional} directly sets a fixed confidence threshold for all categories, and the pseudo labels are obtained when corresponding confidence scores are above such a threshold. However, the fixed threshold mechanism suffers from varying numbers of pseudo labels for different categories, which unavoidably hurts the final segmentation performance. ADVENT \cite{vu2019advent} introduces the category-wise ratio priors on source domain to guide the pseudo label selection. Nevertheless, it still remains challenging to avoid choosing pseudo labels biased towards easy categories.

In this paper, to address the above issues, we propose a category-adaptive domain adaptation approach for semantic segmentation, as illustrated in Fig. \ref{fig:network_figure}. First, adversarial learning is introduced into \textit{style gap bridging mechanism}, in place of widely-used mean squared error (MSE) as an optimization term \cite{li2019high, hou2020source}, since the high dimensional style vector follows such a complex distribution that Gaussian distribution assumption is ill-suited. Further, to balance the probability of chosen pseudo labels for each category, we propose a \textit{category-adaptive threshold method} to construct pseudo labels for SSL. The category-adaptive confidence thresholds are learned for different categories according to their respective contributions.

The main contributions of this paper are summarized as follows: \vspace{-0.3em}
\begin{enumerate}
	\item We propose a \textit{style gap bridging} mechanism based on adversarial learning, which narrows the style-based gap to help alleviate the domain discrepancy. \vspace{-0.3em}
	\item We propose a \textit{category-adaptive threshold} mechanism for pseudo labeling to help SSL on the target domain images.	\vspace{-0.3em}
	\item We conduct a series of experiments on cross-domain segmentation task and verify the effectiveness and superiority of our method.
\end{enumerate}

\vspace{-6mm}
\section{PROPOSED METHOD}
\vspace{-2mm}
\label{sec:format}
\begin{figure*}[!htbp]
	\begin{center}
		\includegraphics[width=0.8\linewidth]{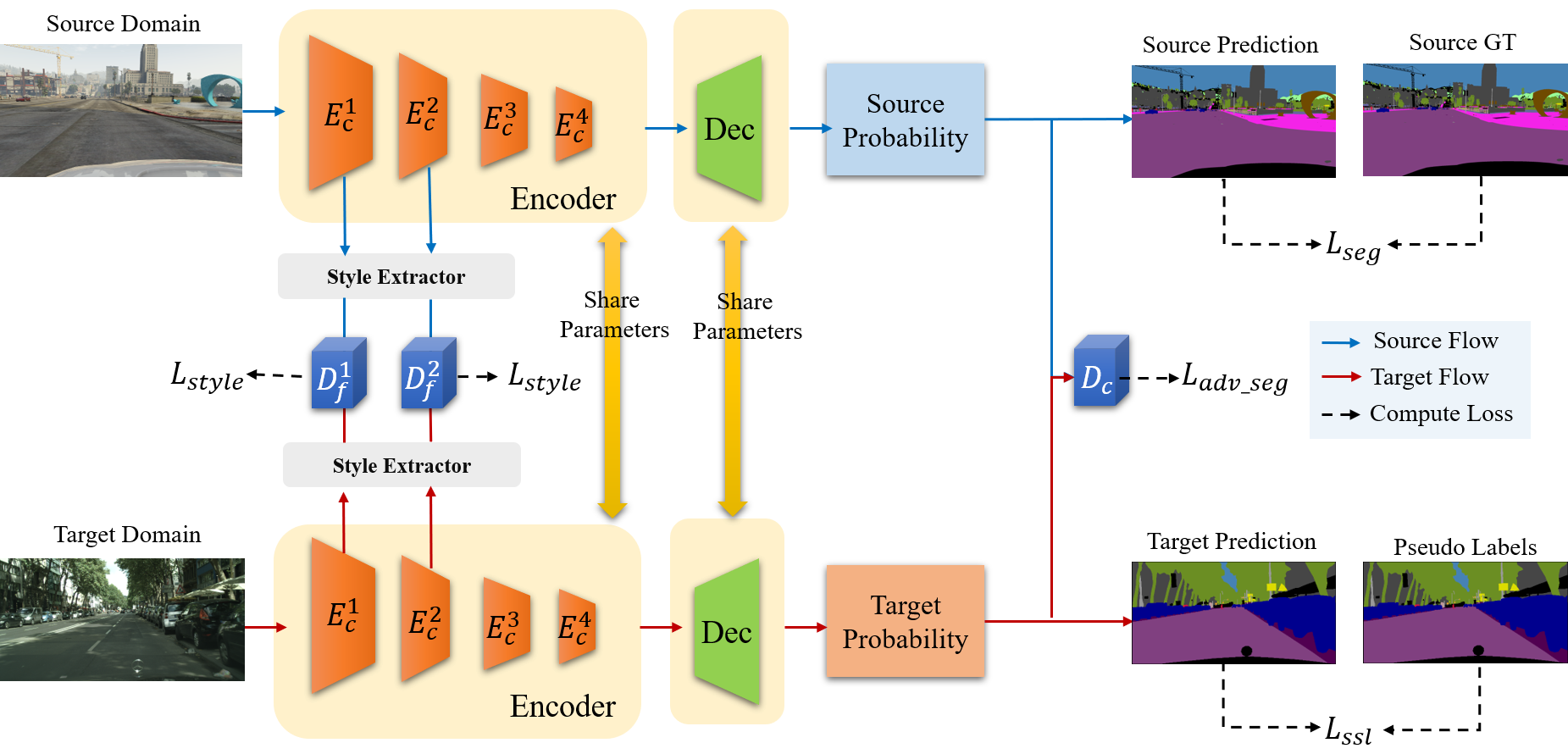}
	\end{center}
	\caption{The framework of our proposed model. Noting that content-based gap is not taken into account, here we take two datasets with regard to urban streetscape for example. The blue flow shows the process of the source domain images, while the red flow shows the process of the target domain images. 
		The encoder module and decoder module are shared for two domain images, where the style extractor is achieved by global average pooling process.  The model is firstly trained on the source domain in a supervised manner. Then it is trained on the target domain with SSL, where the pseudo labels for each category are filtered by the corresponding category-adaptive threshold.}
	\label{framework}
	\vspace{-4mm} 
\end{figure*}
The framework of our proposed model is shown in Fig. \ref{framework}.
Firstly, the model is  trained on both domain data, 
where the style information between both domains is aligned as close as possible.  It is worth noticing that data of the target domain are lack of pixel-wise domain annotations. Then pseudo labels are chosen based on the prediction of pretrained model on target domain. At last, SSL is conducted on the target domain  by virtue of chosen pseudo labels.
\vspace{-5mm}
\subsection{Style Gap Bridging Mechanism}
\label{encoder}

The core of our  encoder is to keep content information, meanwhile, decrease style information as much as possible, since the semantic performance heavily depends on content information. Therefore, it is reasonable to narrow the gaps of style information between source domain images and target domain images. In this paper, without loss of generality, we leverage \textit{global average pooling} as the style extractor in Fig. \ref{framework}, since channel-wise statistics are demonstrated related to style information \cite{huang2017arbitrary}. Previous works \cite{li2019high, hou2020source} usually apply MSE as style constraints, however, MSE performs worse on data with high dimensions and is limited by the linearity and Gaussianity assumptions \cite{lu2013correntropy}. By contrast, adversarial learning is theoretically proved to narrow the gap between two high-dimensional distributions. In practice, with the help of style discriminators (i.e., $D_f^1$ and $D_f^2$), we apply adversarial loss on style information $S_{*n}$ extracted from  2 front sub-encoder modules (i.e., $E_c^1$ and $E_c^2$ in Fig.\ref{framework}), where $*= s/ t$ denotes the source domain / target domain, $n=\{1,2\}$. 
\vspace{-5mm}
\subsection{Pseudo labeling for target domain}
\label{Pseudo labeling}
Here we propose a category-adaptive threshold method for SSL. The idea is based on the hypothesis that the pretrained model's performances on different categories are different because of the uneven  prior distributions of different categories. For example, the  category ``road'' accounts a lot while the  category ``train'' is just the reverse.  Therefore, the confidence threshold should vary among different categories. Based on the clustering method of \cite{zhang2019category} where the threshold is defined by the Euclidean  distance between target features and category centroids, we consider that each intra-category feature makes different contributions to the category centroids because the prediction confidence varies. Consequently, based on the given model's output on the target domain $P_t \in \mathbb{R}^{H_t\times  W_t \times C}$, we firstly define a confidence-weighted target domain-based category centroid $f^l \in \mathbb{R}^C$: 
{\setlength\abovedisplayskip{0.8pt}\setlength\belowdisplayskip{0.8pt}
	\begin{align}
		f^l = \frac{1}{|P^l|}\sum_{h=1}^{H_t}\sum_{w=1}^{W_t}\sum_{c=1}^{C}\hat{y}_t^{hwc}P_t^{hwc},
	\end{align}
}where $P^l$ denotes the collection of prediction confidence of all pixels decided as $l$-th category, $|P^l|$ denotes the cardinality of $P^l$. $\hat{y}_t^{hwc}=\mathbbm{1}_{[c=\mathop{\arg\max}\limits_{c'}p_T^{hwc'}]}$ , and $\mathbbm{1}$ is the binary indicator function.

Given $f^l$ in each category, our threshold is based on the entropy distance. The entropy of prediction vector at the  $h$th row and $w$th column $P_t^{hw} \in \mathbb{R}^C$ is:
{\setlength\abovedisplayskip{0.8pt}\setlength\belowdisplayskip{1pt}
	\begin{align}
		\label{entropy}
		E(P_t^{hw}) = -\sum_{i=1}^{C}P_t^{hwc}\log P_t^{hwc}.
	\end{align}
}

The entropy of category centroid $f^l$, namely $E(f^l)$ is similar with Equation \eqref{entropy}. Intuitively,  $E(P_t^{hw})$ decreases as the max confidence in $P_t^{hw}$ increases,  consequently we choose entropy-based threshold. Here we defined an indicator variable $m_t^{hwc}$ to decide whether the prediction on current position is chosen as available pseudo labels:
{\setlength\abovedisplayskip{1pt}\setlength\belowdisplayskip{1pt}
	\begin{align}
		\label{modified_pl}
		m_t^{hwc} = \mathbbm{1}_{[E(P_t^{hw})<E(f^l)-\vartriangle]},
	\end{align}
} 
\vspace{-3mm}

\noindent where $\vartriangle$ is a manually fixed hyperparameter to control the threshold for each category. When $\vartriangle$ increases, the number of available pseudo labels decreases while the model will have higher prediction confidence  and vice versa. 
\vspace{-3mm}
\subsection{Loss Functions}
\label{loss}
\vspace{-1mm}
As  mentioned above, the training process includes two phases: domain adaptation training  and SSL. Domain adaptation training process utilizes the following three losses:

\textbf{Segmentation Loss.} Here cross entropy function is applied to penalize the error between prediction $\hat{y}_s \in \mathbb{R}^{H_s \times W_s \times C}$ and one-hot ground truth $y_s \in \mathbb{R}^{H_s \times W_s \times C}$:
{\setlength\abovedisplayskip{1pt}\setlength\belowdisplayskip{1pt}
	\begin{align}
		\mathcal{L}_{seg} = -\frac{1}{H_s \times W_s}\sum_{h=1}^{H_s}\sum_{w=1}^{W_s}\sum_{c=1}^{C}y_s^{hwc}\log\hat{y}_s^{hwc}.
	\end{align}
}

\textbf{Output-based Domain Adaptation Loss.} Consistent with BDL~\cite{li2019bidirectional}, we also leverage the original GAN loss introduced by Goodfellow \cite{goodfellow2014generative}  as $\mathcal{L}_{adv\_seg}$ to achieve domain adaptation on models' output between the source domain and the target domain, which is achieved by means of the segmentation discriminator $D_c$.

\textbf{Style Loss.}  To help the encoder module $E_c$ extract style-independent features, $\mathcal{L}_{style}$  also utilizes the original GAN loss \cite{goodfellow2014generative} to force the style information on the source domain $S_{sn}$ close that on the target domain $S_{tn}$.

The loss function during domain adaptation training is summarized as follows
{\setlength\abovedisplayskip{1pt}\setlength\belowdisplayskip{1pt}
	\begin{align}
		\label{loss_function}
		\mathcal{L} = \lambda_{seg} \mathcal{L}_{seg}  + \lambda_{adv\_seg} \mathcal{L}_{adv\_seg} +\lambda_{style} \mathcal{L}_{style},
\end{align}}where $\lambda$s play a trade-off among these three terms.

During the SSL process, similar with $\mathcal{L}_{seg}$, \textbf{Self-supervised Loss} $\mathcal{L}_{ssl}$ also utilizes cross entropy function to make the prediction on the target domain $\hat{y}_t \in \mathbb{R}^{H_t\times W_t\times C }$  as close as possible to  pseudo labels $y_t \in \mathbb{R}^{H_t\times W_t\times C}$:
{\setlength\abovedisplayskip{1pt}\setlength\belowdisplayskip{1pt}
\begin{align}
	\mathcal{L}_{ssl} = -\frac{1}{H_t\times W_t}\sum_{h=1}^{H_t}\sum_{w=1}^{W_t}\sum_{c=1}^{C}m_t^{hwc}\hat{y}_t^{hwc}logP_t^{hwc}.
\end{align}
}

\vspace{-5mm}
\section{Experimental Results}
Here we evaluate our model on ``GTA5  to Cityscapes'' task. 
\vspace{-2mm}
\subsection{Datasets}
\label{dataset}
\vspace{-2mm}

\textbf{GTA5 \cite{richter2016playing}} includes 24966 synthetic images collected from the game engine. GTA5 have 19-category pixel-accurate annotations compatible with target domain Cityscapes \cite{cordts2016cityscapes}.

\textbf{Cityscapes \cite{cordts2016cityscapes}} is collected from streetscapes in 50 different Germany cities includes training set with 2975 images, validation set with 500 images, testing set with 1525 images. The former two sets contain pixel-wise semantic label maps, while the annotations of testing set are missing. To validate the performance of our model, during testing phase, we use validation set instead of testing set.
\vspace{-3mm}
\subsection{Network Architectures and Implementation Details.}
\label{network}
The whole framework of our model is shown in Fig. \ref{framework}. The encoder module follows DeepLab V2 \cite{chen2017deeplab} using ResNet101 \cite{he2016deep} as backbone. The parameters are tuned based on weights pretrained  on ImageNet \cite{krizhevsky2017imagenet}. The discriminator $D_c$ for output-based domain adaptation applies PatchGAN \cite{DBLP:journals/corr/IsolaZZE16} to output a 16x downsampled confidence probability map relative to the input semantic segmentation map. The style discriminator $D_f$ also utilizes PatchGAN \cite{DBLP:journals/corr/IsolaZZE16}, but it applies four 1-D convolutional layers with kernel size of 4. All modules are parameter-shared except the style discriminators (i.e., $D_f^1$ and $D_f^2$), and segmentation discriminator $D_c$.  

Note that the all game-synthetic source domain images (GTA5 datasets) are firstly translated by CycleGAN \cite{zhu2017unpaired} module of BDL model \cite{li2019bidirectional}.
SGD optimizer with $momentum = 0.9$ is used to train encoder $E_c$ and decoder modules, where encoder $E_c$ adopts learning rate $lr = 2.5\times 10^{-4}$, the decoder adopts $lr = 2.5 \times 10^{-3}$. For style discriminator $D_f$ and segmentation discriminator, Adam optimizer is utilized with $\beta=(0.9,0.99)$ and $lr = 1\times 10^{-4}$. In addition, ``poly'' policy for learning rate update with $max step=250,000$ and $power=0.9$ is introduced to encoder $E_c$ and decoder. $\lambda_{seg}, \lambda_{adv\_seg}, \lambda_{style}$ in Equation \eqref{loss_function} are set $1$, $1\times 10^{-3}$, $1\times 10^{-3}$,  respectively. Style discriminator $D_f$ and segmentation discriminator $D_c$ utilize exponential decay policy to update $lr$, where  $decay\_rate=0.1$, $decay\_steps=50000$. Two rounds of SSL are applied in our experiments.

\vspace{-5mm}
\subsection{Results}
\label{results}
\vspace{-1mm}
\textbf{Quantitative results:} The results of different related baselines are shown in Table \ref{tab:comparison_gta5}. Consistent with previous work, mIoU metric on 19 specific categories is adopted, where the best result on each category is highlighted in bold. Our model has a gain of 1.7 in overall mIoU rather than the state-of-the-art BDL. In addition, compared to another category-balanced SSL model CBST, our model brings +3.2\% mIoU improvement, which demonstrates the superiority of our proposed pseudo labeling method.

\begin{table*}[!htp]
	\scriptsize
	\centering
	\caption{Comparison among different methods for ``GTA5  to Cityscapes''}
	\label{tab:comparison_gta5}
	\setlength{\tabcolsep}{3pt}
	\resizebox{0.95\textwidth}{!}{
		\begin{tabular}{ccccccccccccccccccccc}
			\hline
			\multicolumn{21}{c}{{ GTA5 $\rightarrow$ Cityscapes}} \\
			\hline
			{Method}  & \rotatebox{90}{road}  & \rotatebox{90}{sidewalk} &\rotatebox{90}{building} & \rotatebox{90}{wall} & \rotatebox{90}{fence} & \rotatebox{90}{pole} & \rotatebox{90}{t-light} & \rotatebox{90}{t-sign} & \rotatebox{90}{vegetation} & \rotatebox{90}{terrain} & \rotatebox{90}{sky} & \rotatebox{90}{person} & \rotatebox{90}{rider} & \rotatebox{90}{car} & \rotatebox{90}{truck} & \rotatebox{90}{bus} & \rotatebox{90}{train} & \rotatebox{90}{motorbike} & \rotatebox{90}{bicycle} & mIoU\\
			\hline
			CBST\cite{zou2018unsupervised} &89.6 &\bf{58.9} &78.5 &33.0 &22.3 &\bf{41.4} &\bf{48.2} &39.2 &83.6 &24.3 &65.4 &49.3 &20.2 &83.3 &39.0 &48.6 &\bf{12.5} &20.3 &35.3 &47.0\\
			\hline
			Cycada \cite{hoffman2018cycada} & 86.7 & 35.6 & 80.1 & 19.8 & 17.5 & {{38.0}} & {39.9} & {\bf{41.5}} & 82.7 & 27.9 & 73.6 & {\bf{64.9}} & 19 & 65.0 & 12.0 & 28.6 & 4.5 & 31.1 & {42.0} & 42.7 \\
			\hline
			ADVENT \cite{vu2019advent} & 87.6 & 21.4 & 82.0 & 34.8 & 26.2 & 28.5 & 35.6 & 23.0 & 84.5 & 35.1 & 76.2 & 58.6 & 30.7 & 84.8 &34.2 & 43.4 & 0.4 & 28.4 & 35.2 & 44.8\\
			\hline
			DCAN \cite{wu2018dcan} & 85.0 & 30.8 & 81.3 & 25.8 & 21.2 & 22.2 & 25.4 & 26.6 & 83.4 & 36.7 & 76.2 & 58.9 & 24.9 & 80.7 & 29.5 & 42.9 & 2.5 & 26.9 & 11.6 & 41.7 \\
			\hline
			CLAN \cite{luo2019taking} & 87.0 & 27.1 & 79.6 & 27.3 & 23.3 & 28.3 & 35.5 & 24.2 & 83.6 & 27.4 & 74.2 & 58.6 & 28.0 & 76.2 & 33.1 & 36.7 & 6.7 & {\bf{31.9}} & 31.4 & 43.2 \\
			\hline
			BDL \cite{li2019bidirectional} & {91.0} & {44.7} & {84.2} & {34.6} & {\bf{27.6}} & 30.2 & 36.0 & 36.0 & \textbf{85.0} & {\bf{43.6}} & {83.0} & 58.6 & {31.6} & {83.3} & {35.3} & {49.7} & 3.3 & 28.8 & 35.6 & {48.5} \\
			\hline
			Ours & \bf{91.7} & {{51.1}} & {\bf{85.0}} & \bf{38.7} & {26.7} & 32.1 & 38.1 & 34.6 & 84.3 & {38.6} & \bf{84.9} & 60.7 & \bf{32.8} & \bf{85.2} & {\bf{41.9}} & {\bf{49.8}} & 2.8 & 28.5 & \bf{45.0} & \bf{50.2} \\ \hline
		\end{tabular}
	}
	\vspace{-1em}
\end{table*}

\vspace{-3mm}
\begin{figure}[!h]
	\centering
	\begin{subfigure}[b]{0.24\linewidth}
		\centering
		\includegraphics[width=\linewidth]{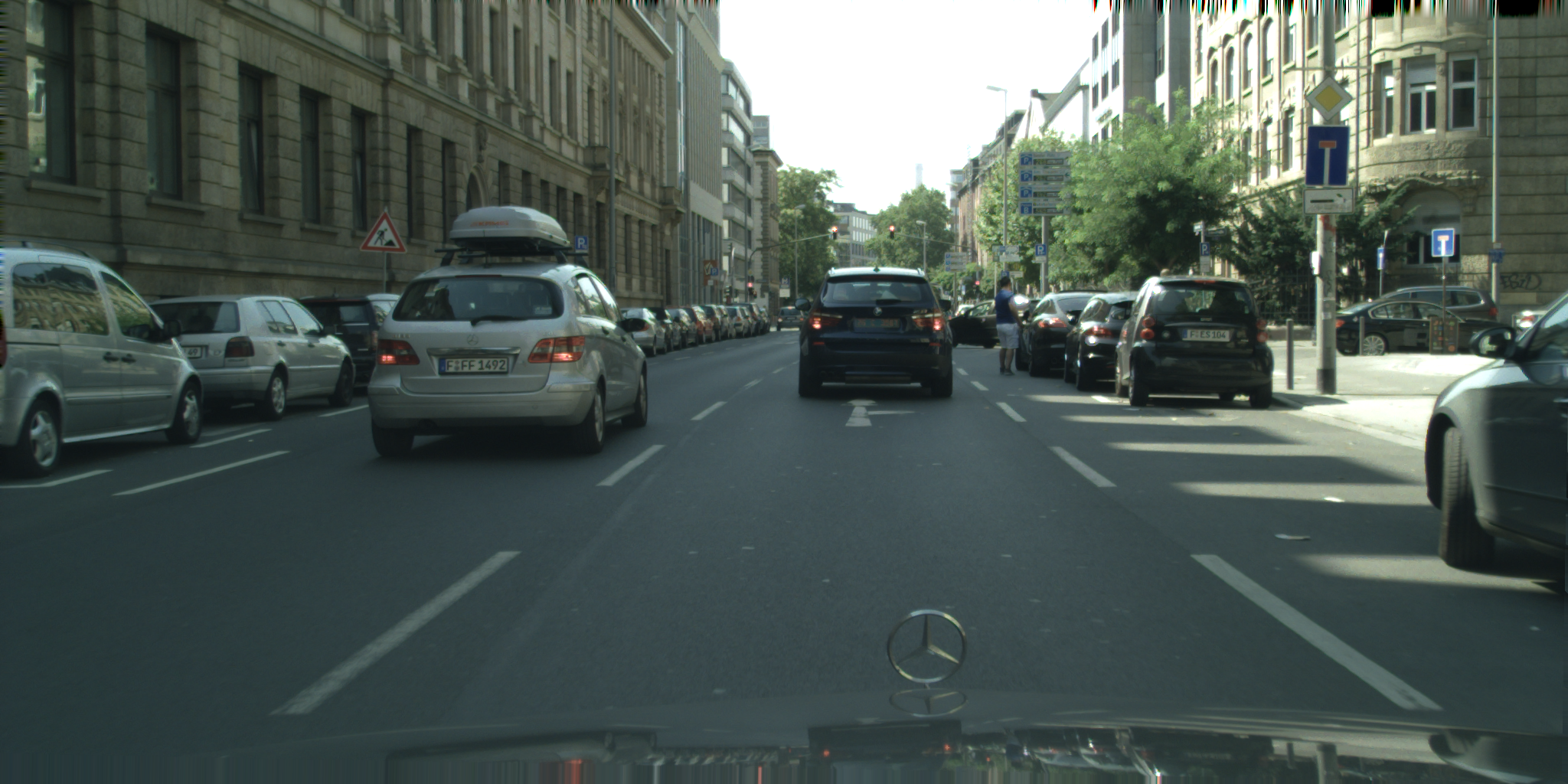}
	\end{subfigure}
	\begin{subfigure}[b]{0.24\linewidth}
		\centering
		\includegraphics[width=\linewidth]{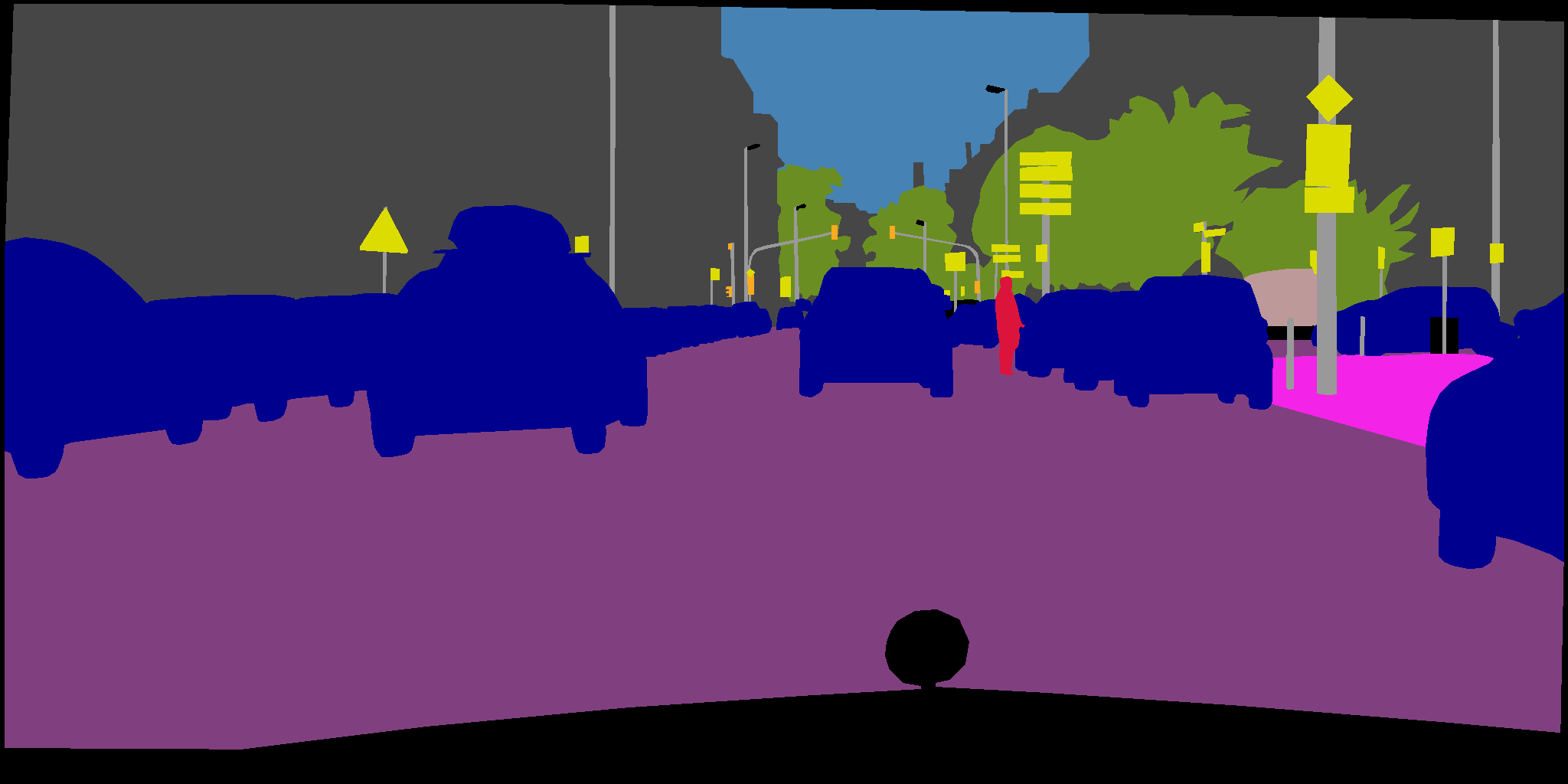}
	\end{subfigure}
	\begin{subfigure}[b]{0.24\linewidth}
		\centering
		\includegraphics[width=\linewidth]{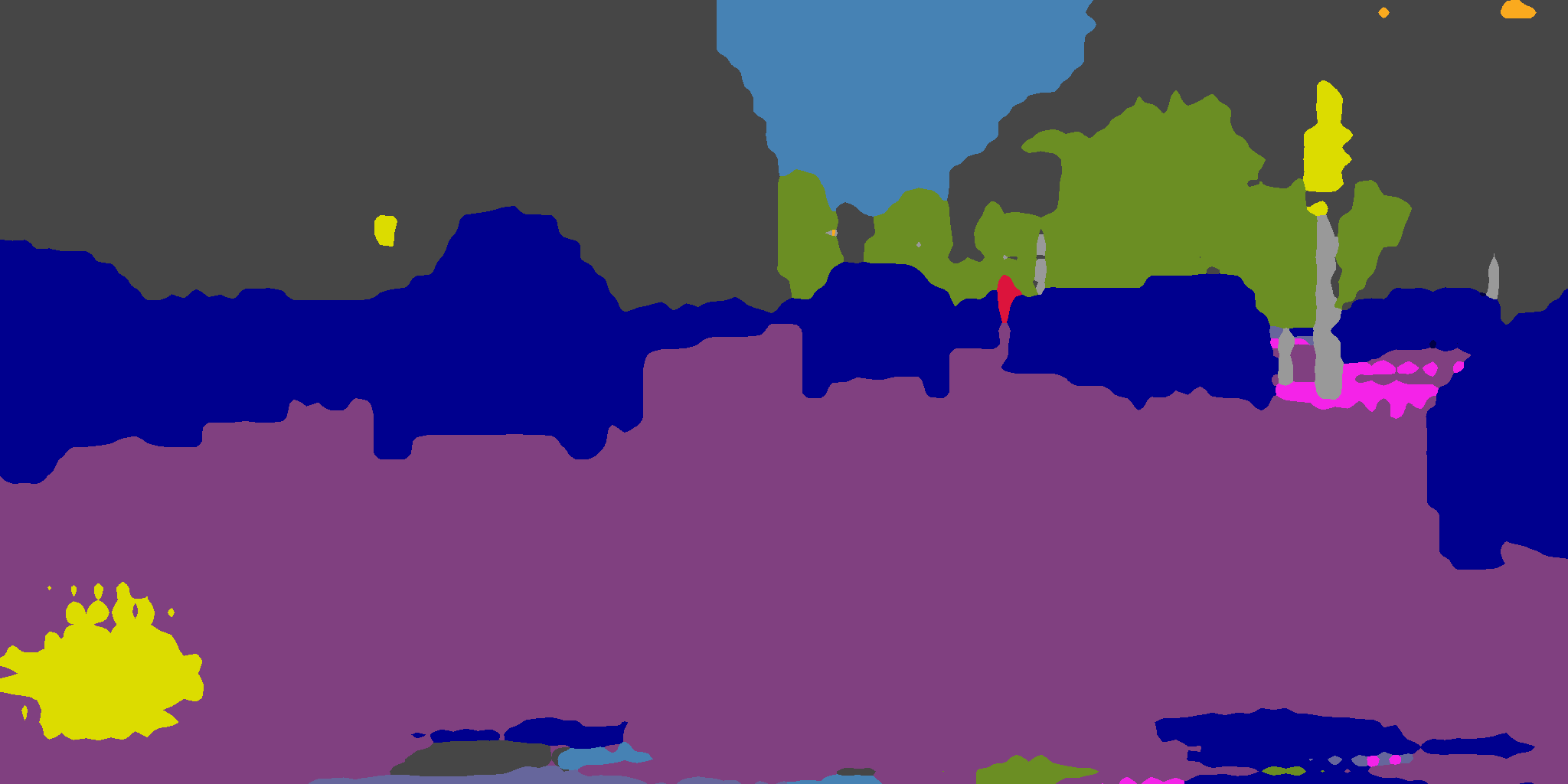}
	\end{subfigure}
	\begin{subfigure}[b]{0.24\linewidth}
		\centering
		\includegraphics[width=\linewidth]{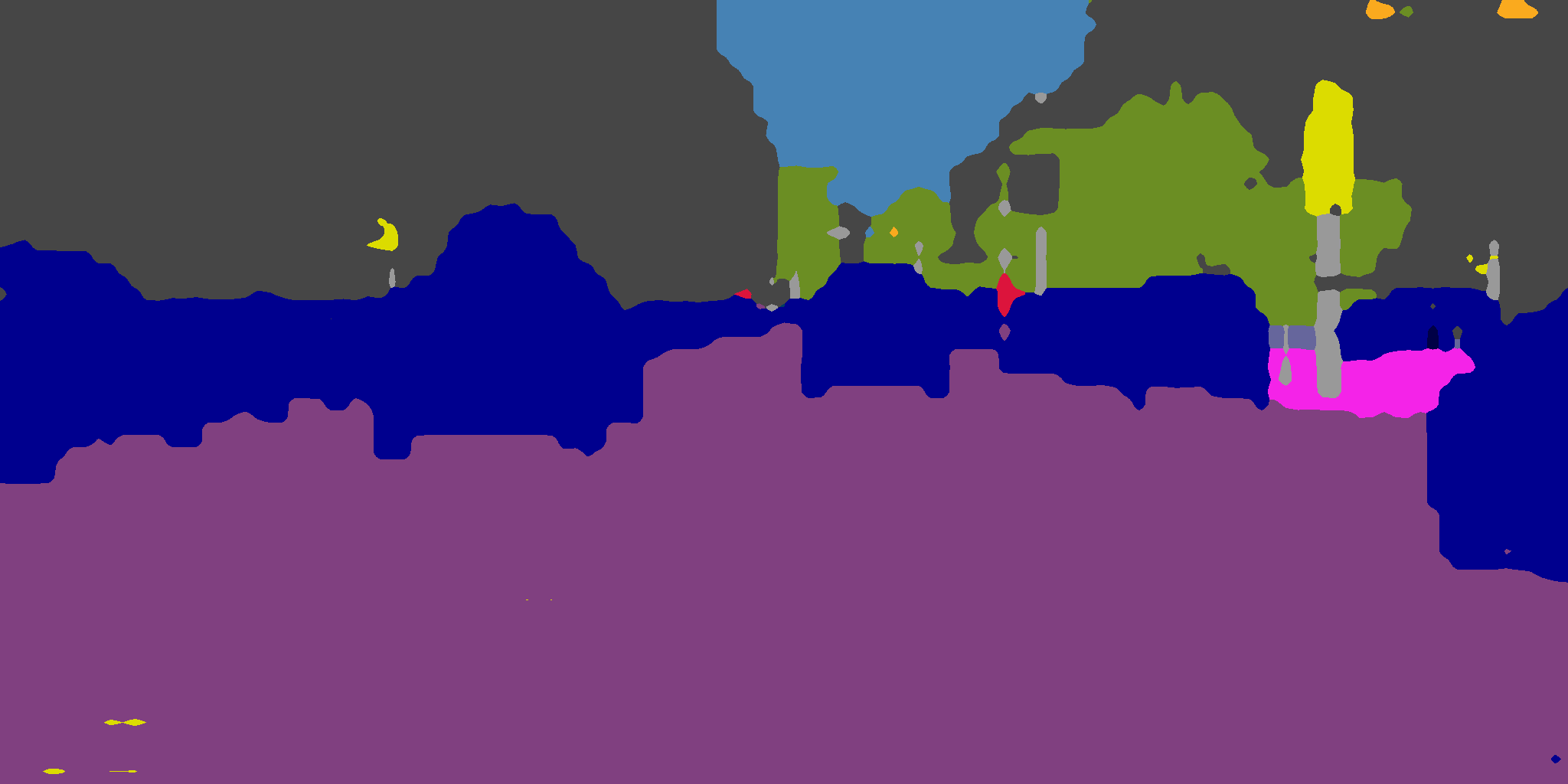}
	\end{subfigure}
	\\
	\vspace{.05cm}
	\begin{subfigure}[b]{0.24\linewidth}
		\centering
		\includegraphics[width=\linewidth]{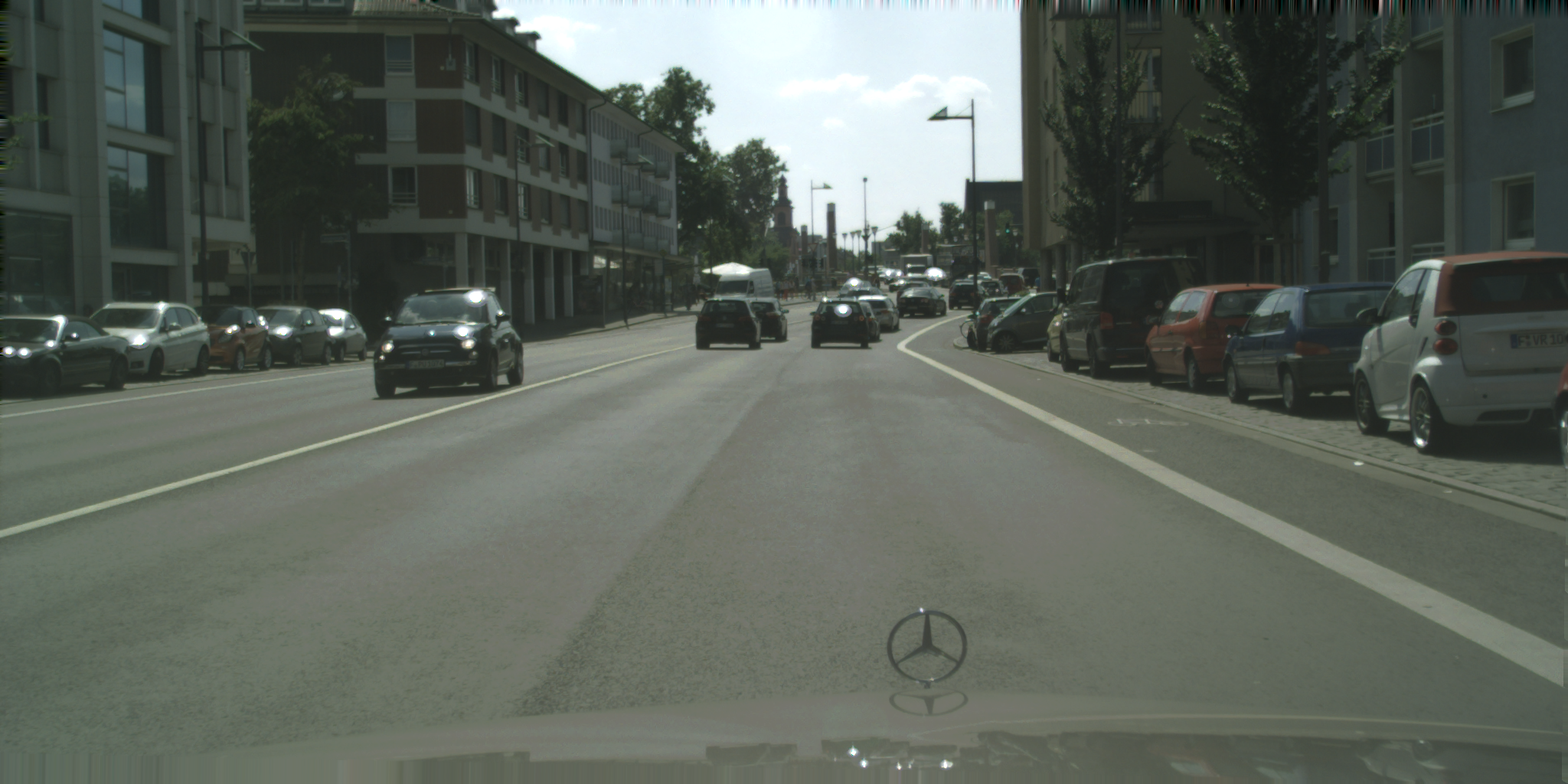}
	\end{subfigure}
	\begin{subfigure}[b]{0.24\linewidth}
		\centering
		\includegraphics[width=\linewidth]{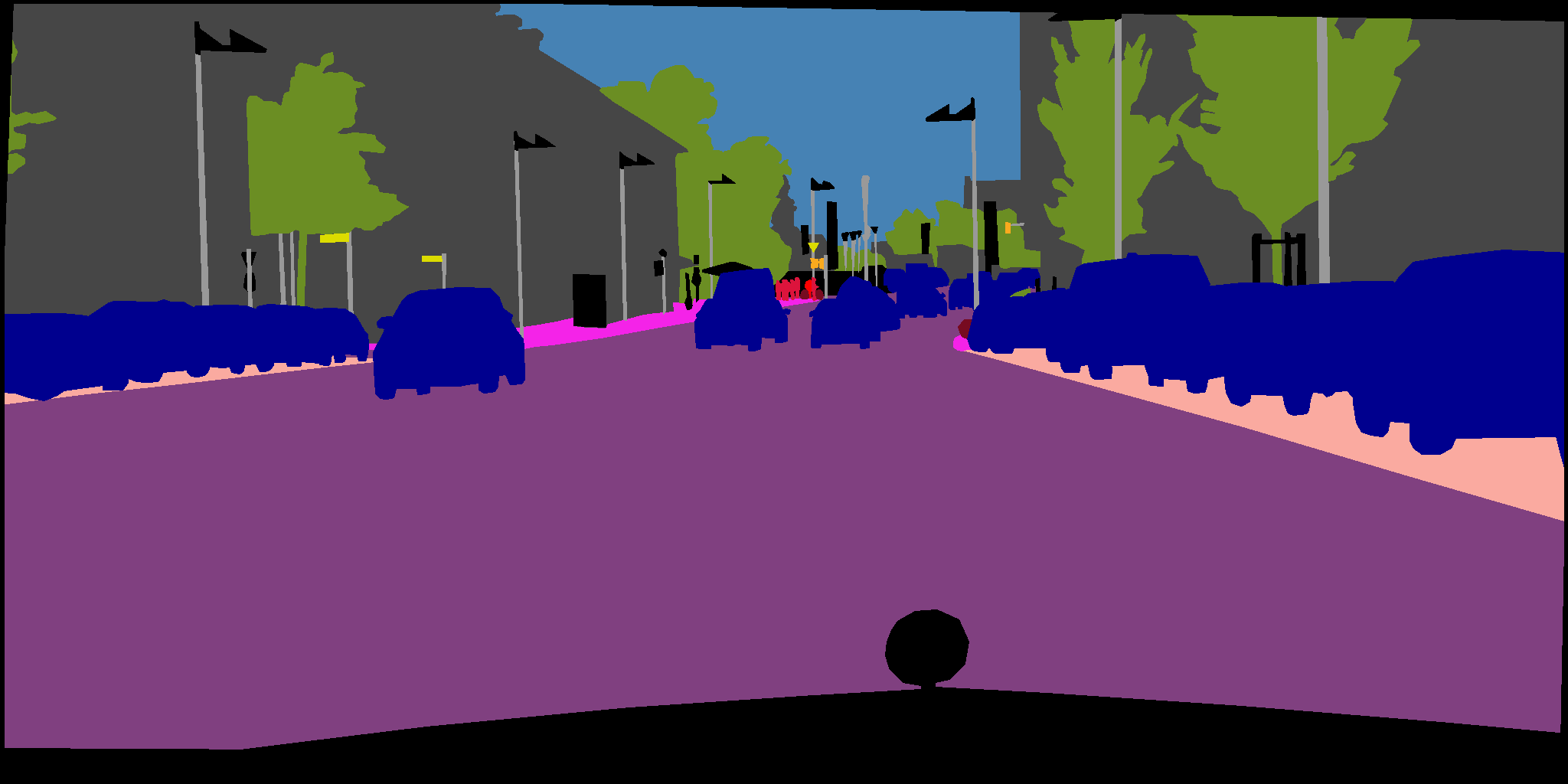}
	\end{subfigure}
	\begin{subfigure}[b]{0.24\linewidth}
		\centering
		\includegraphics[width=\linewidth]{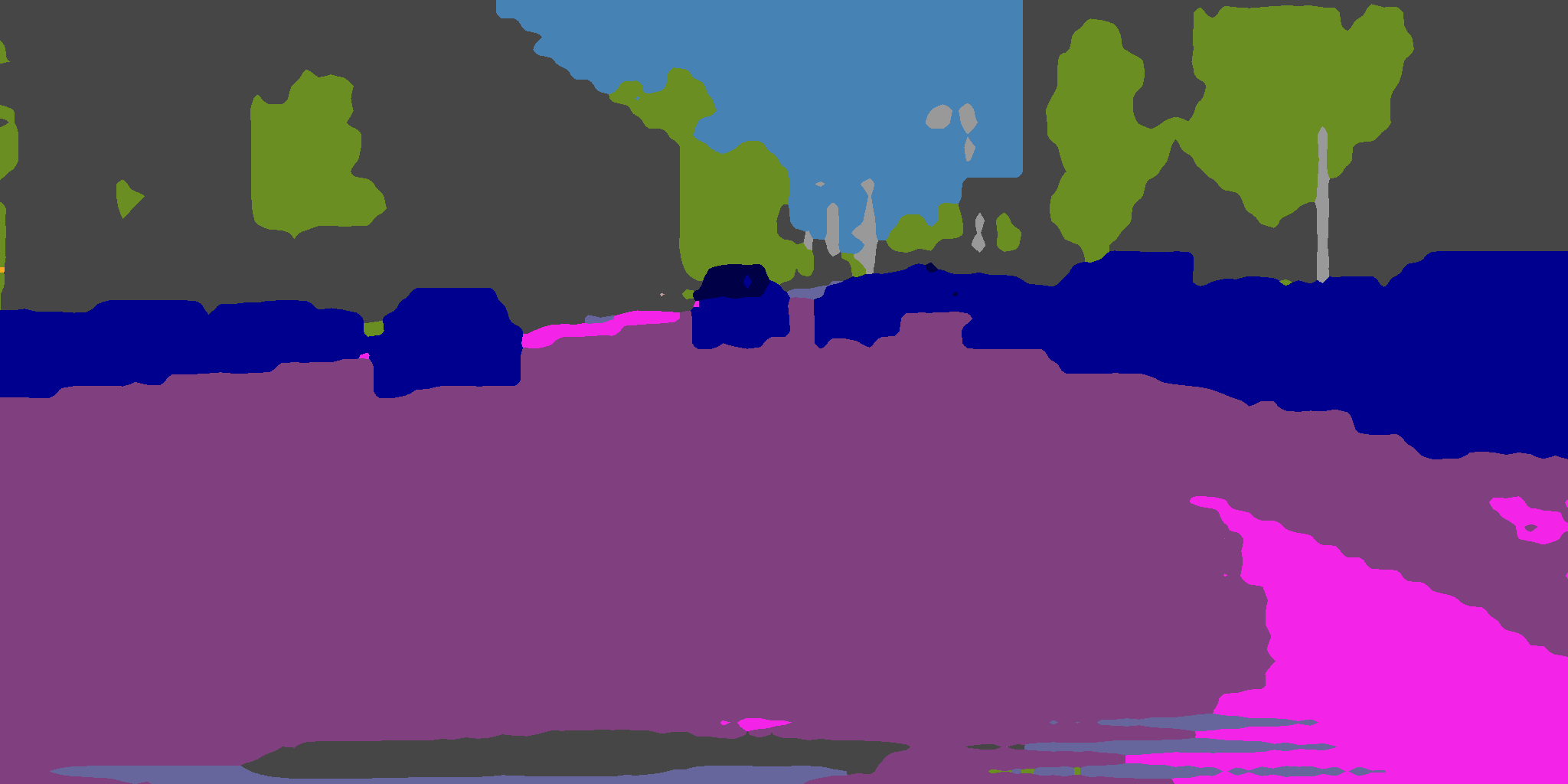}
	\end{subfigure}
	\begin{subfigure}[b]{0.24\linewidth}
		\centering
		\includegraphics[width=\linewidth]{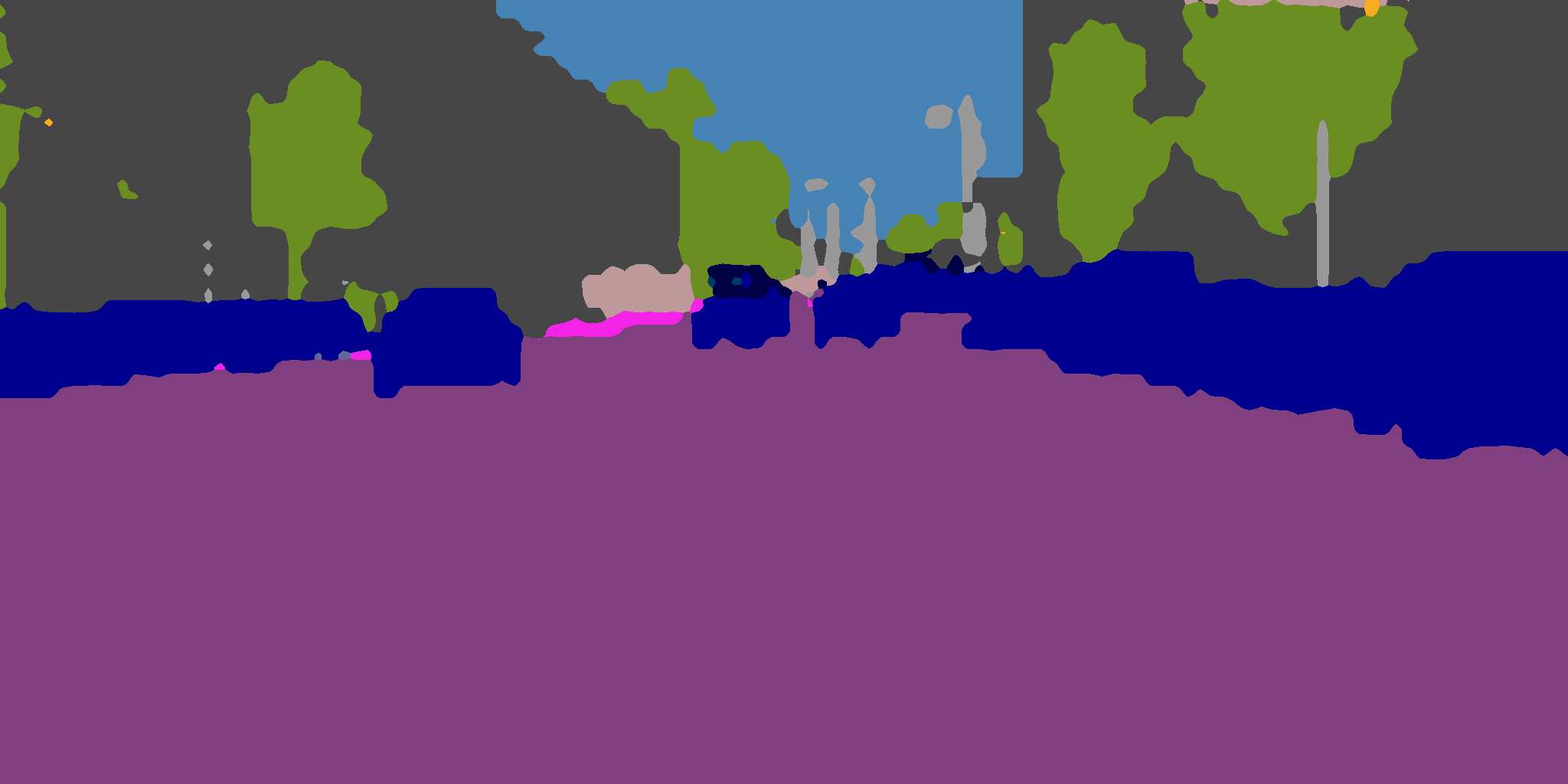}
	\end{subfigure}
	\\
	\vspace{.05cm}
	\begin{subfigure}[b]{0.24\linewidth}
		\centering
		\includegraphics[width=\linewidth]{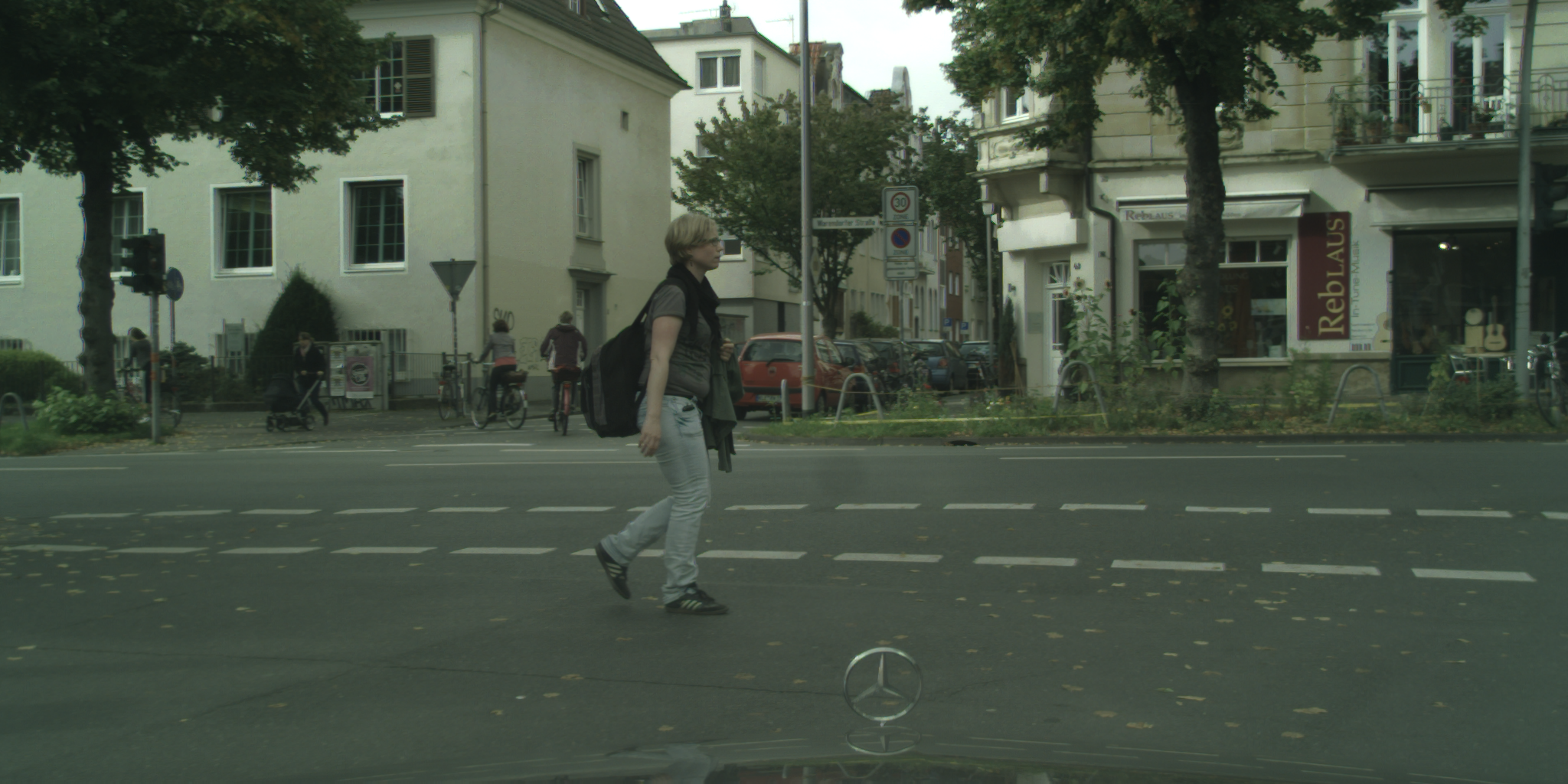}
		\caption{\label{fig:a}Image}
	\end{subfigure}
	\begin{subfigure}[b]{0.24\linewidth}
		\centering
		\includegraphics[width=\linewidth]{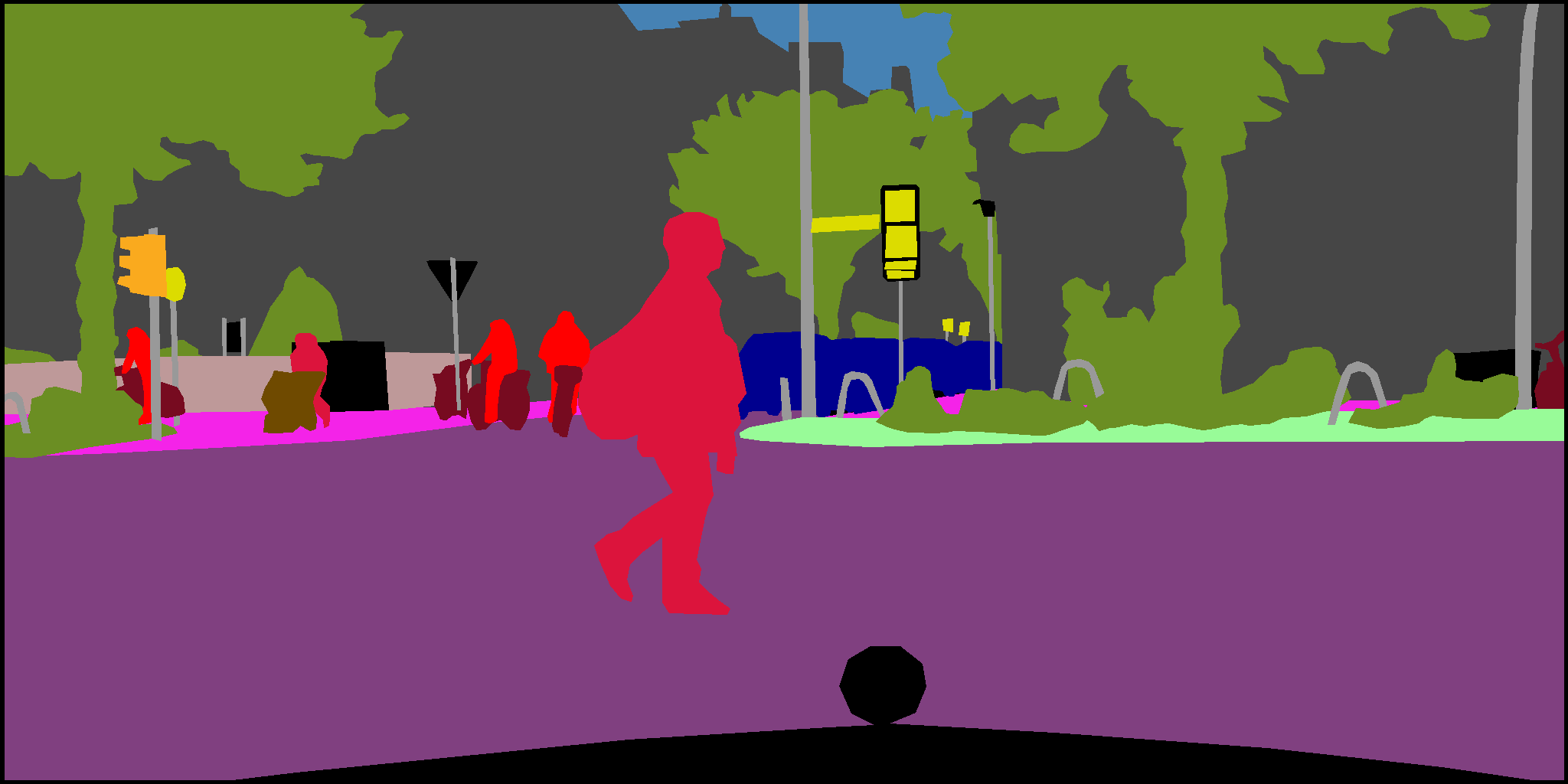}
		\caption{\label{fig:b}GT}
	\end{subfigure}
	\begin{subfigure}[b]{0.24\linewidth}
		\centering
		\includegraphics[width=\linewidth]{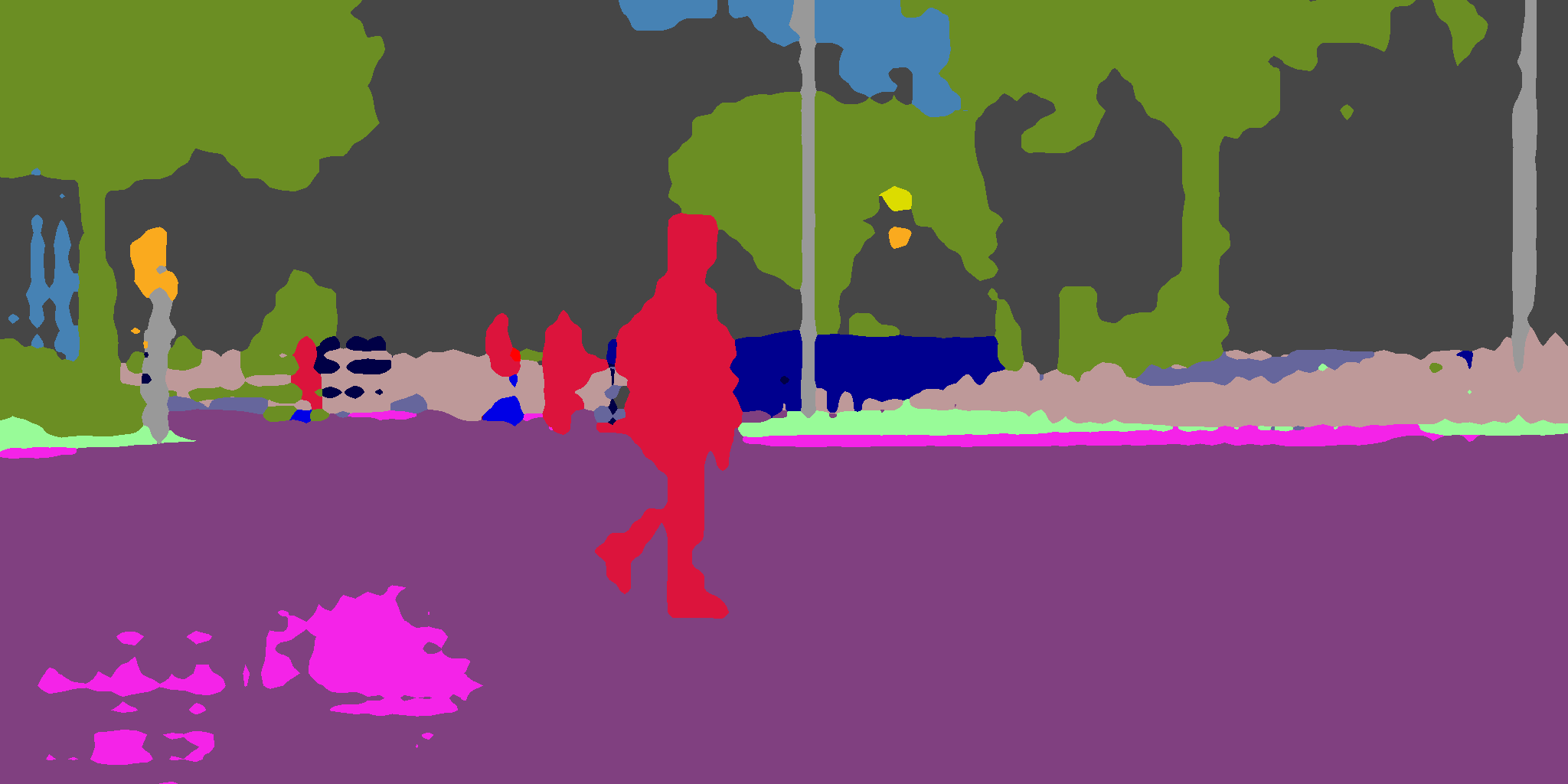}
		\caption{\label{fig:c}BDL(gta2cs)}
	\end{subfigure}
	\begin{subfigure}[b]{0.24\linewidth}
		\centering
		\includegraphics[width=\linewidth]{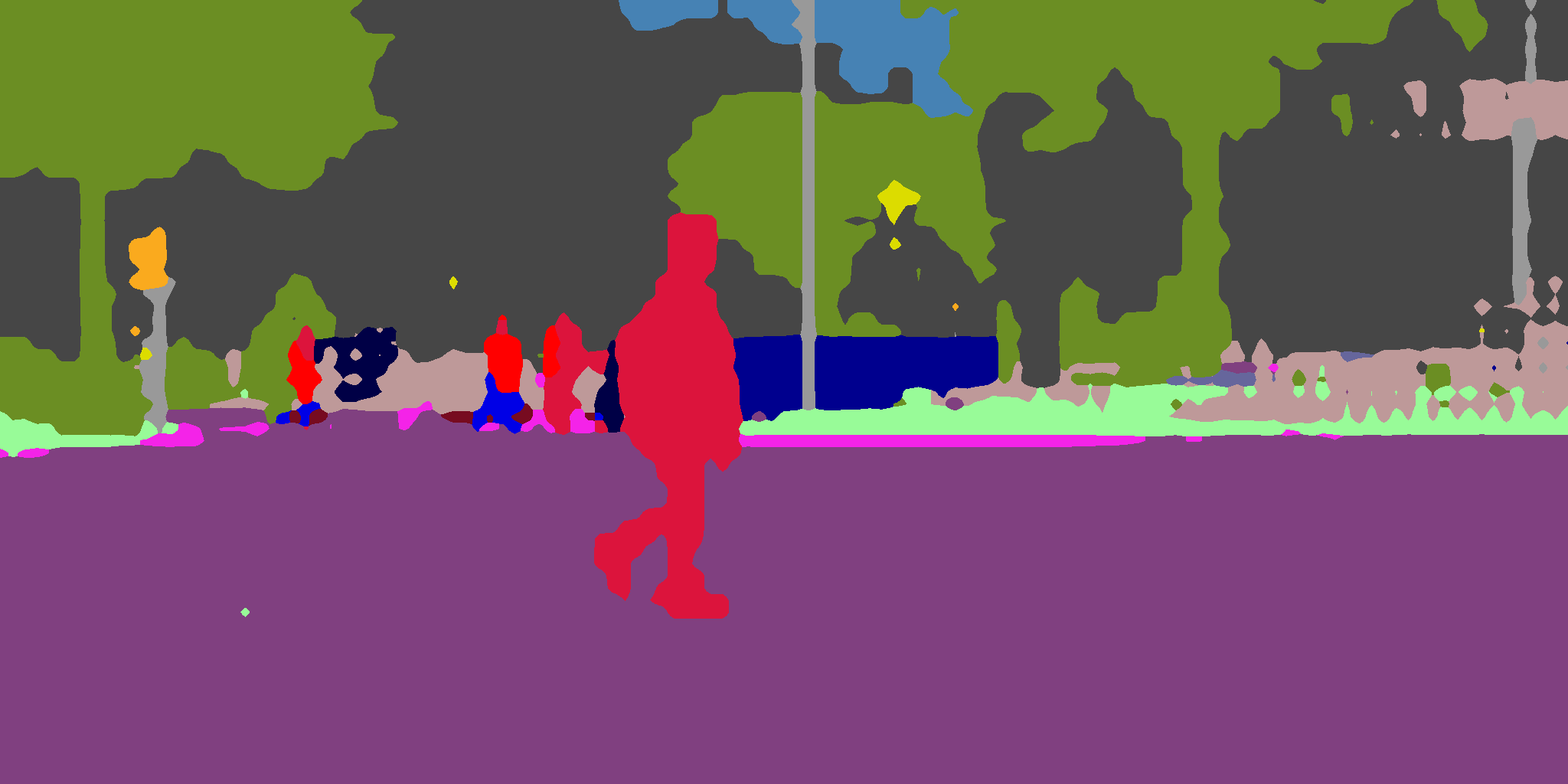}
		\caption{\label{fig:d}Ours(gta2cs)}
	\end{subfigure}
	\caption{\label{fig}Qualitative comparisons. From left to right: (a) Original Cityscape images, (b) Ground truth, (c) BDL on ``GTA5 to Cityscapes'', (d) Ours on ``GTA5 to Cityscapes''.}
	\label{qualitative}
\end{figure}
\vspace{-2mm}
\begin{table}[!hbt]
	\caption{Ablation study on SSL and style constraints.}
	\centering
	\scriptsize
	\resizebox{0.6\linewidth}{!}{
		\begin{tabular}{cc}
			\hline
			\multicolumn{2}{ c }{{ GTA5 $\rightarrow$ Cityscapes}} \\
			\hline
			model & mIoU \\
			\hline
			original & 44.6 \\
			\hline
			original + adv & 45.5\\
			\hline
			original + adv + SSL once   & 48.5\\
			\hline
			original + adv + SSL twice  & 50.2\\
			\hline
		\end{tabular}
		\label{tab:ablation}
		\vspace{-1em}
	}
\end{table}
\vspace{-3mm}
\textbf{Ablation Study:}
Our main contributions consist of a novel pseudo labeling mechanism for SSL and adversarial learning based style gap bridging mechanism. Table \ref{tab:ablation} illustrates the influence of each part, where ``original'' denotes the model without these two parts, ``adv'' implies style constraints in an adversarial manner and ``SSL once'' and ``SSL twice'' refer to  using SSL once and twice, respectively. It can be seen that style constraints help improve the performance with a gain of 0.9 on mIoU, which demonstrates that adversarial learning indeed narrows the style gap between two domains. In addition, SSL is also helpful to boost performance since ``SSL once'' brings a gain of 3.0 and ``SSL twice'' achieves 50.2,  which is 4.7\% superior to that without SSL module. 

In addition, to compare different style gap bridging mechanisms, we also conduct check experiments (no SSLs) with the same settings. The results are  shown in Table \ref{tab:style}, where ``mean \& std" refers to  the channel-wise means and standard deviations of given features. Note that our method does not exploit second order statistics compared with the left two methods but outperforms Gram matrix based and ``mean \& std" based methods with a gain of 0.8 and 0.4, which demonstrates the superiority of adversarial learning as the style gap bridging mechanism compared to MSE constraints. 
\vspace{-2mm}
\begin{table}[!hbt]
	\caption{Comparison on style gap bridging mechanisms}
	\centering
	\scriptsize
	\label{style_modeling}
	\resizebox{0.75\linewidth}{!}{
		\begin{tabular}{c|c|c}
			\hline
			\multicolumn{1}{p{2cm}|}{\centering style gap bridging mechanism} & \multirow{1}[3]{*}{style modeling} & \multirow{1}[3]{*}{mIoU} \\
			\hline
			\multirow{2}{*}{MSE} & Gram matrix & 44.7 \\
			\cline{2-3}
			& mean \& std & 45.1 \\
			\hline
			adversarial learning  & mean (Ours) & 45.5 \\
			\hline

		\end{tabular}
		\label{tab:style}
		\vspace{-1em}
	}
\end{table}

\vspace{-1mm}
\textbf{Qualitative results:}
Some segmentation examples are shown in Fig. \ref{qualitative}. It can be clearly observed that our method  makes less visually obvious prediction errors than BDL,
\vspace{-5mm}
\section{Conclusion}
\label{sec:conclusion}
\vspace{-3mm}
In this paper, we proposed a style gap bridging mechanism and category-adaptive threshold method for SSL on cross-domain semantic segmentation task. The former utilizes adversarial training to narrow the gaps of style information. The latter makes the use of prior semantic distributions to dynamically choose thresholds for self-supervised training on the target domain images, instead of applying fixed thresholds. A series of experiments have shown the effectiveness and superiority of our proposed model.	
\vspace{-4mm}
\section{Acknowledgement}
\vspace{-2mm}
This work is supported in part by the National Key R\&D Program of China under Grant 2018YFA0701601, and in part by the fellowship of China National Postdoctoral Program for Innovative Talents (BX20200194).
	
    \bibliographystyle{IEEEtran}
    \bibliography{refs.bib}

\begin{thebibliography}{10}
\providecommand{\url}[1]{#1}
\csname url@samestyle\endcsname
\providecommand{\newblock}{\relax}
\providecommand{\bibinfo}[2]{#2}
\providecommand{\BIBentrySTDinterwordspacing}{\spaceskip=0pt\relax}
\providecommand{\BIBentryALTinterwordstretchfactor}{4}
\providecommand{\BIBentryALTinterwordspacing}{\spaceskip=\fontdimen2\font plus
\BIBentryALTinterwordstretchfactor\fontdimen3\font minus
  \fontdimen4\font\relax}
\providecommand{\BIBforeignlanguage}[2]{{%
\expandafter\ifx\csname l@#1\endcsname\relax
\typeout{** WARNING: IEEEtran.bst: No hyphenation pattern has been}%
\typeout{** loaded for the language `#1'. Using the pattern for}%
\typeout{** the default language instead.}%
\else
\language=\csname l@#1\endcsname
\fi
#2}}
\providecommand{\BIBdecl}{\relax}
\BIBdecl

\bibitem{zeiler2011adaptive}
M.~D. Zeiler, G.~W. Taylor, and R.~Fergus, ``Adaptive deconvolutional networks
  for mid and high level feature learning,'' in \emph{2011 International
  Conference on Computer Vision}.\hskip 1em plus 0.5em minus 0.4em\relax IEEE,
  2011, pp. 2018--2025.

\bibitem{gatys2015neural}
L.~A. Gatys, A.~S. Ecker, and M.~Bethge, ``A neural algorithm of artistic
  style,'' \emph{arXiv preprint arXiv:1508.06576}, 2015.

\bibitem{huang2017arbitrary}
X.~Huang and S.~Belongie, ``Arbitrary style transfer in real-time with adaptive
  instance normalization,'' in \emph{Proceedings of the IEEE International
  Conference on Computer Vision}, 2017, pp. 1501--1510.

\bibitem{zou2018unsupervised}
Y.~Zou, Z.~Yu, B.~Vijaya~Kumar, and J.~Wang, ``Unsupervised domain adaptation
  for semantic segmentation via class-balanced self-training,'' in
  \emph{Proceedings of the European conference on computer vision (ECCV)},
  2018, pp. 289--305.

\bibitem{li2019bidirectional}
Y.~Li, L.~Yuan, and N.~Vasconcelos, ``Bidirectional learning for domain
  adaptation of semantic segmentation,'' in \emph{Proceedings of the IEEE
  Conference on Computer Vision and Pattern Recognition}, 2019, pp. 6936--6945.

\bibitem{vu2019advent}
T.-H. Vu, H.~Jain, M.~Bucher, M.~Cord, and P.~P{\'e}rez, ``Advent: Adversarial
  entropy minimization for domain adaptation in semantic segmentation,'' in
  \emph{Proceedings of the IEEE conference on computer vision and pattern
  recognition}, 2019, pp. 2517--2526.

\bibitem{li2019high}
M.~Li, C.~Ye, and W.~Li, ``High-resolution network for photorealistic style
  transfer,'' \emph{arXiv preprint arXiv:1904.11617}, 2019.

\bibitem{hou2020source}
Y.~Hou and L.~Zheng, ``Source free domain adaptation with image translation,''
  \emph{arXiv preprint arXiv:2008.07514}, 2020.

\bibitem{lu2013correntropy}
C.~Lu, J.~Tang, M.~Lin, L.~Lin, S.~Yan, and Z.~Lin, ``Correntropy induced l2
  graph for robust subspace clustering,'' in \emph{Proceedings of the IEEE
  international conference on computer vision}, 2013, pp. 1801--1808.

\bibitem{zhang2019category}
Q.~Zhang, J.~Zhang, W.~Liu, and D.~Tao, ``Category anchor-guided unsupervised
  domain adaptation for semantic segmentation,'' in \emph{Advances in Neural
  Information Processing Systems}, 2019, pp. 435--445.

\bibitem{goodfellow2014generative}
I.~Goodfellow, J.~Pouget-Abadie, M.~Mirza, B.~Xu, D.~Warde-Farley, S.~Ozair,
  A.~Courville, and Y.~Bengio, ``Generative adversarial nets,'' in
  \emph{Advances in neural information processing systems}, 2014, pp.
  2672--2680.

\bibitem{richter2016playing}
S.~R. Richter, V.~Vineet, S.~Roth, and V.~Koltun, ``Playing for data: Ground
  truth from computer games,'' in \emph{European conference on computer
  vision}.\hskip 1em plus 0.5em minus 0.4em\relax Springer, 2016, pp. 102--118.

\bibitem{cordts2016cityscapes}
M.~Cordts, M.~Omran, S.~Ramos, T.~Rehfeld, M.~Enzweiler, R.~Benenson,
  U.~Franke, S.~Roth, and B.~Schiele, ``The cityscapes dataset for semantic
  urban scene understanding,'' in \emph{Proceedings of the IEEE conference on
  computer vision and pattern recognition}, 2016, pp. 3213--3223.

\bibitem{chen2017deeplab}
L.-C. Chen, G.~Papandreou, I.~Kokkinos, K.~Murphy, and A.~L. Yuille, ``Deeplab:
  Semantic image segmentation with deep convolutional nets, atrous convolution,
  and fully connected crfs,'' \emph{IEEE transactions on pattern analysis and
  machine intelligence}, vol.~40, no.~4, pp. 834--848, 2017.

\bibitem{he2016deep}
K.~He, X.~Zhang, S.~Ren, and J.~Sun, ``Deep residual learning for image
  recognition,'' in \emph{Proceedings of the IEEE conference on computer vision
  and pattern recognition}, 2016, pp. 770--778.

\bibitem{krizhevsky2017imagenet}
A.~Krizhevsky, I.~Sutskever, and G.~E. Hinton, ``Imagenet classification with
  deep convolutional neural networks,'' \emph{Communications of the ACM},
  vol.~60, no.~6, pp. 84--90, 2017.

\bibitem{DBLP:journals/corr/IsolaZZE16}
\BIBentryALTinterwordspacing
P.~Isola, J.~Zhu, T.~Zhou, and A.~A. Efros, ``Image-to-image translation with
  conditional adversarial networks,'' \emph{CoRR}, vol. abs/1611.07004, 2016.
  [Online]. Available: \url{http://arxiv.org/abs/1611.07004}
\BIBentrySTDinterwordspacing

\bibitem{zhu2017unpaired}
J.-Y. Zhu, T.~Park, P.~Isola, and A.~A. Efros, ``Unpaired image-to-image
  translation using cycle-consistent adversarial networks,'' in
  \emph{Proceedings of the IEEE international conference on computer vision},
  2017, pp. 2223--2232.

\bibitem{hoffman2018cycada}
J.~Hoffman, E.~Tzeng, T.~Park, J.-Y. Zhu, P.~Isola, K.~Saenko, A.~Efros, and
  T.~Darrell, ``Cycada: Cycle-consistent adversarial domain adaptation,'' in
  \emph{International conference on machine learning}.\hskip 1em plus 0.5em
  minus 0.4em\relax PMLR, 2018, pp. 1989--1998.

\bibitem{wu2018dcan}
Z.~Wu, X.~Han, Y.-L. Lin, M.~Gokhan~Uzunbas, T.~Goldstein, S.~Nam~Lim, and
  L.~S. Davis, ``Dcan: Dual channel-wise alignment networks for unsupervised
  scene adaptation,'' in \emph{Proceedings of the European Conference on
  Computer Vision (ECCV)}, 2018, pp. 518--534.

\bibitem{luo2019taking}
Y.~Luo, L.~Zheng, T.~Guan, J.~Yu, and Y.~Yang, ``Taking a closer look at domain
  shift: Category-level adversaries for semantics consistent domain
  adaptation,'' in \emph{Proceedings of the IEEE Conference on Computer Vision
  and Pattern Recognition}, 2019, pp. 2507--2516.

\end{thebibliography}

\end{document}